Article

# Application of Deep Reinforcement Learning to UAV Swarming for Ground Surveillance

Raúl Arranz, David Carramiñana, Gonzalo de Miguel, Juan A. Besada and Ana M. Bernardos *

Information Processing and Telecommunications Center, Universidad Politécnica de Madrid, ETSI Telecomunicación, Av. Complutense 30, 28040 Madrid, Spain; raul.arranz@upm.es (R.A.); d.carraminana@upm.es (D.C.); gonzalo.demiguel@upm.es (G.d.M.); juanalberto.besada@upm.es (J.A.B.)
* Correspondence: anamaria.bernardos@upm.es

**Abstract:** This paper summarizes in depth the state of the art of aerial swarms, covering both classical and new reinforcement-learning-based approaches for their management. Then, it proposes a hybrid AI system, integrating deep reinforcement learning in a multi-agent centralized swarm architecture. The proposed system is tailored to perform surveillance of a specific area, searching and tracking ground targets, for security and law enforcement applications. The swarm is governed by a central swarm controller responsible for distributing different search and tracking tasks among the cooperating UAVs. Each UAV agent is then controlled by a collection of cooperative sub-agents, whose behaviors have been trained using different deep reinforcement learning models, tailored for the different task types proposed by the swarm controller. More specifically, proximal policy optimization (PPO) algorithms were used to train the agents' behavior. In addition, several metrics to assess the performance of the swarm in this application were defined. The results obtained through simulation show that our system searches the operation area effectively, acquires the targets in a reasonable time, and is capable of tracking them continuously and consistently.

**Keywords:** artificial intelligence; swarm; drones; search; tracking; obstacle avoidance; centralized

## 1. Introduction

Swarms of unmanned aerial vehicles (UAVs) have been the focus of many research studies in recent years due to their relevance and adaptability in real-life applications. A swarm consists of a group of UAVs which perform coordinated operations to perform a target task. There exists a huge number of strategies for swarming depending on the service case, the properties of the vehicles in use, the application domain, the coordination and planning algorithms employed, etc. Some applications studied in the literature are surveillance [1], search and rescue [2], payload transportation [3], reconnaissance and mapping [4], and public communications [5].

After a complete survey of swarming methods, this paper will focus in solving a surveillance problem using a swarm of UAVs with on-board sensors. The surveillance function of the swarm includes both searching and tracking ground targets, while avoiding collisions, both with (dynamic) obstacles and between UAVs in the swarm. Potential uses of such a surveillance swarm could be border control, search and rescue, or critical infrastructure protection. The system proposed in this paper uses hybrid AI approaches, integrating a deterministic controller managing the high-level mission of the swarm with deep reinforcement learning algorithms to obtain a behavior model for planning the path of each individual drone within the swarm.

The proposed system aims to solve the aforementioned ground surveillance problem, taking into account the particular environment and problem requirements and dynamics, and modeling it more realistically than other approaches in the literature. A distinct feature of our solution is that the proposed model takes into account the need to avoid static and dynamic obstacles in the environment while taking into account the UAVs' and target's dynamics, sensor coverage, etc. Another novel feature is the use of hybrid AI to ease the training process of our system. Following the proposed approach, a complex task is divided into several individual models that can be trained individually. Then, the whole system behavior is obtained by combining these models using a series of rules.

The rest of the paper is organized as follows. First, a survey of the state of the art of swarm-based surveillance is included in Section 2. Then, Section 3 describes our swarming system proposal, requirements, architecture, and training process. This system has been evaluated using simulation-based experiments, whose results are included in Section 4. Finally, Section 5 concludes the paper and describes a collection of future lines for research.

## 2. State of the Art

UAV swarming use cases and control algorithms have been widely studied in the literature. In fact, multiple planning and control methods can be used, resulting in different swarming architectures. Of particular interest to organize the discussion is the well-known swarming conceptual architecture proposed in [6] and depicted in Figure 1. In this generic architecture, the swarming process is decomposed into five layers, each targeting an individual problem. It is important to notice that the architecture is conceptual, and actual implementations may jointly implement functions at different layers. A more detailed description of its layers follows.

1. Mission-planning layer: High-level layer which is responsible for the evaluation, planning, and assignment of tasks to individual UAVs and UAV formations (or clusters), and generates decision data for the path-planning layer.
2. Path-planning layer: Mid-level layer that manages the tasks and generates corresponding task planning paths for each UAV or formation based on the decision data.
3. Formation-control/collision-avoidance layer: This performs task coordination between several nearby UAVs according to the path information, and implements automatic obstacle avoidance and formation control.
4. Communication layer: This conducts network communication according to the interactive information generated by the control layer to implement the necessary information sharing between individuals.
5. Application layer: This will feed back the corresponding environment information to the mission-planning layer according to different application scenarios, closing the loop and enabling adaptation to the dynamic scenario.

This section will first analyze different swarm management organizations. Then, swarming methods and algorithms will be discussed, using Figure 1 as a conceptual reference for the state-of-the-art analysis, especially addressing the mission-planning, path-planning, and collision-avoidance layers. A particular focus on the state of the art will be in reinforcement learning methods, which will be applied in our proposed swarming system. The section continues by summarizing the survey and it ends by providing a general introduction on reinforcement-learning-based methods, especially important for the rest of the paper.

### 2.1. Swarm Management Organization

A swarm of UAVs can be internally organized in several ways. There are centralized swarms, in which there is one swarm controller (SC) in full command of the UAVs, and decentralized ones, in which UAVs have decision-making capabilities. In decentralized architectures, we can further distinguish hierarchical organizations from fully distributed ones. Next, we will describe this taxonomy of architectures [7–9], which are schematically depicted in Figure 2.

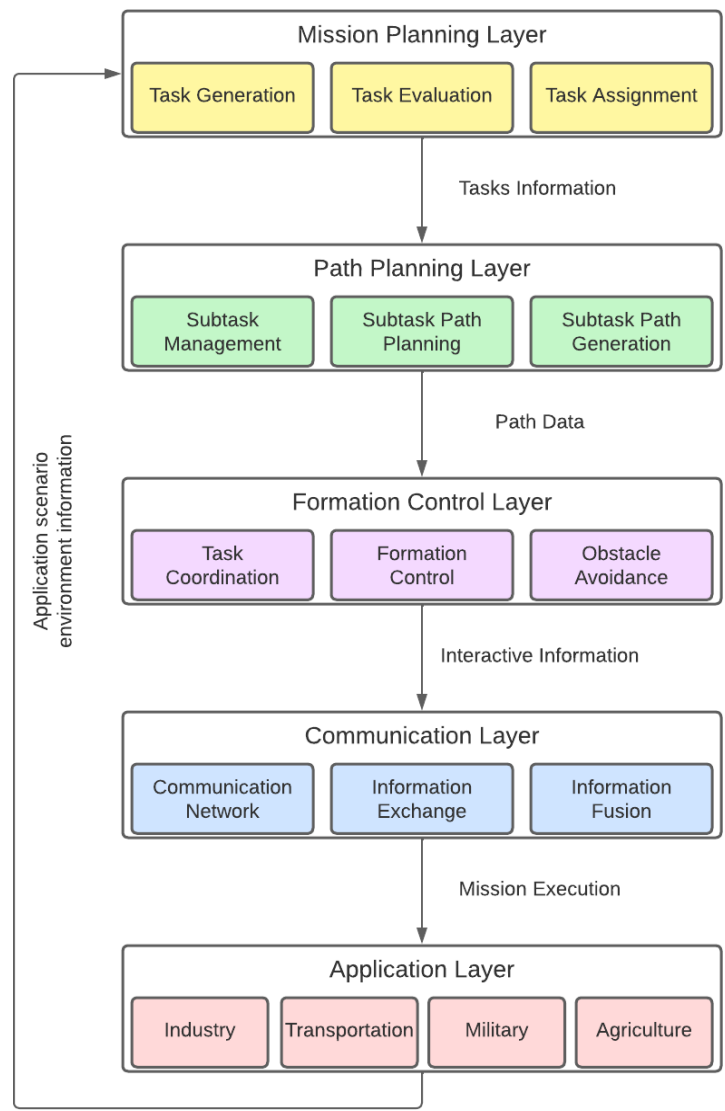

**Figure 1.** Swarming process layers and relationship between them.

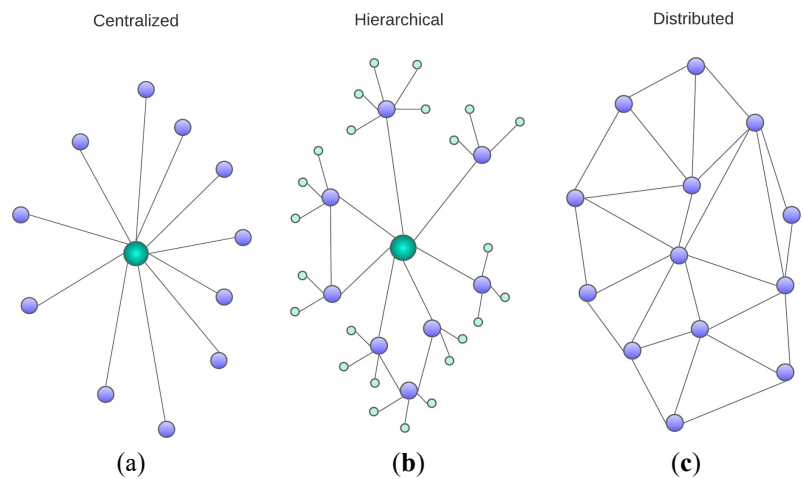

**Figure 2.** Control structures. (**a**) Centralized structure. (**b**) Hierarchical structure. (**c**) Distributed structure.

In centralized control structures (depicted in Figure 2a), a single platform directly manages the whole swarm, meaning that it performs any calculations needed to execute control tasks. The platform can be a vehicle of the swarm or a ground control station (GCS). This effectively makes the centralized architecture the simplest form of control structure. Under this architecture, high-level mission decisions are taken on a mission control center or swarm controller (especially those related to the task assignment in the mission-planning layer). Then, different solutions distribute intermediate functions related to flight task scheduling and path planning so that sometimes they are pre-calculated in the mission control center, and sometimes the UAV controllers have the intelligence to autonomously take decisions in real time pertaining to those layers. Typically, decisions that must be taken by the UAVs autonomously during their operation, even in centralized architectures, are those related to lower-level control tasks (especially those related to the formation-control/collision-avoidance layer description). A disadvantage of centralized systems is that they rely heavily on the existence of safe communication channels, since the control decisions are taken by the central platform and must be transmitted to the followers, who ultimately execute them. This leads to a high amount of communications network traffic and hinders scalability.

Decentralized control structures allow for increased swarm autonomy if the computations are performed using on-board flight controllers. This decentralized approach is more in line with the swarm behavior concept stemming from natural systems, where a group behavior emerges by the interactions of the individual UAVs, and this behavior cannot be simplified to the mere aggregation of its constituents. An advantage of decentralized systems in general is that they are scalable and less reliant on communications stability.

As already mentioned, inside decentralized structures, two categories can be distinguished: hierarchical and fully distributed. Hierarchical systems can be considered as an extension of centralized systems in which some decision autonomy is given to edge platforms. As such, they are sometimes called semi-centralized systems. In hierarchical schemes, the chain-of-command structure takes the form of a tree (see Figure 2b) in which one or several leaders are each in charge of a group of platforms, called followers, which are constrained by the indications of the leader but have a certain degree of autonomy.

Meanwhile, in fully distributed architectures (shown in Figure 2c) there is no leader in charge of the swarm. Instead, all the agents are autonomous and with an equivalent level of intelligence. In this case, the decisions are distributed among all the UAVs. Distributed systems, thus, benefit from higher levels of autonomy, and consequently enhanced fault tolerance. Other advantages that distributed control provides are better scalability and the ability to leverage parallelization.

*2.2. Mission-Planning Layer*

Let us define a collection of concepts to specify the mission-planning layer:

- Each of the *n* UAVs available will be denoted as $U_i$.
- A swarm is the set of all UAVs available: $S = \{U_1, U_2, \ldots, U_n\}$.
- Each task to be performed by the swarm is denoted $T_i$.
- Tasks are not independent, and there are a collection of dependencies $D_{i,j} = \{T_i, T_j\}$, meaning that in order to start executing $T_j$, $T_i$ must have been completed. Here, $T_i$ will be a predecessor task of $T_j$.
- $T = \{T_1, T_2, \ldots, T_k\}$ is the set of all tasks to be performed.
- $D = \{D_{i,j}\}$ is the set of all dependencies.
- The mission $M = \{T, D\}$ is defined as the set of all tasks and dependencies.

The aim of the mission-planning layer is to schedule the mission of the swarm, $M$, in such a way that the total mission cost is minimized. The assignment process produces a mapping (or an assignment) from the set of tasks to the set of UAVs, which specifies which task is assigned to which UAV at each time. In general, the mission might change dynamically, as new needs arrive or new information is available at the swarm level. The mission cost might be modeled as the sum of the costs associated to execute all tasks.

Different types of tasks may have quite different costs attached, but in general we may decompose the cost into a traveling cost and a task completion cost. The former is the cost for a UAV to travel to the task location from its current location, while the latter is the cost for the UAV to carry out the task.

Quite often, to solve this problem, two planning levels are combined: global task assignment and local task scheduling. In centralized planning architectures, these levels are respectively performed by two types of agents: a central swarm controller (SC) and the UAV agent. The SC may be understood as the logical interface between a swarm operator and the swarm drones. It usually runs in the ground station and it aims to (1) decompose the operator high-level missions into multiple executable tasks, (2) assign those tasks to UAVs, and (3) monitor the status of the swarm drones and mission completion. On the other hand, each UAV agent schedules and executes assigned tasks, focusing on the flight planning and control of each UAV.

The SC needs to know all the information about the swarm and its operational environment to make task assignment decisions. For each UAV, some interesting information to be periodically collected might be its current location, its battery/fuel level, the status of completion of the tasks, estimates of its position, and expected battery/fuel level after all assigned tasks are completed by the UAV, etc. Furthermore, every time a UAV finishes a task it should inform the SC to enable it to monitor the mission execution completion status and to assess the potential assignation of additional tasks. Through time, the SC needs to decide which tasks are ready to be assigned. A task is said to be ready if it does not have any predecessor or all its predecessors have been updated to completed. The SC then monitors the status of each mission and assigns tasks to UAVs as they become ready. For each ready task, the SC will select the UAV to perform it, for instance, according to the following steps (which perform a local optimization):

- Compute the cost the task will involve for each UAV. The SC will use its latest knowledge about the swarm to estimate the cost of finishing it. This cost may be different for different UAVs. For a UAV which does not have scheduled tasks, the cost is the traveling cost from its current position plus the task's completion cost. Conversely, if the UAV has already scheduled tasks, the task cost is calculated considering the estimated final location after completing the current tasks.
- Assign the task to a UAV. A potential candidate would be the UAV with the smallest cost.

Other SC task assignation policies might be used, but in general they should follow the same approach: model somehow the cost for each UAV to execute the task, and assign it taking into account individual costs. Also, the benefit of performing the task for the overall swarm could be modeled and taken into account, especially if we are in a resource-constrained situation, where some tasks might be discarded or delayed.

The task assignment process sends tasks to the UAVs' control agents, but it does not necessarily specify how this task should be scheduled for the UAV, which might control this decision, and permute the task execution order. For the UAV agent, there are several possible scheduling policies, some potential examples (with different characteristics regarding their local or global optimality, calculation cost, etc.) follow.

- First come, first served: The new task will be executed after all earlier tasks are completed.
- Insertion-based policy: If there are no scheduled tasks on the UAV, the new task will start execution immediately. Otherwise, the UAV will schedule the new task to optimize its own overall cost. If before the new task is assigned the UAV must perform $N$ sorted tasks $\{T_1, T_2, \ldots, T_N\}$, a way to recalculate the plan would be placing the new task $\hat{T}$ in each position from 1 to $N + 1$ and computing every cost of performing all the tasks, i.e., the cost of ordered sets $\{\hat{T}, T_1, T_2, \ldots, T_N\}$, $\{T_1, \hat{T}, T_2, \ldots, T_N\}$, $\ldots, \{T_1, T_2, \ldots, T_N, \hat{T}\}$. The policy will choose the position of the new task that minimizes the cost.
- Use of classical solutions to the traveling salesman problem (TSP) [10]. If task execution costs are assumed to be independent of the task order, traveling costs drive the difference between the different scheduling processes, and then the application of those solutions are direct.
- Adaptive policy: This policy simply uses other policies to calculate task costs and picks the policy that produces the minimum cost each time the UAV needs to perform a task scheduling.

These scheduling processes at the SC and UAVs' agents should collaboratively check the feasibility of the task lists, i.e., it may happen that the UAV's battery/fuel is not enough to complete the whole process, or the tasks cannot be completed in a reasonable time. In such cases, tasks will be removed until the individual UAV's plans become feasible, leading potentially to incomplete mission fulfillment.

In general, the mission-planning problem can be understood as an optimization one, where existing optimization techniques can be applied. In fact, there exist an extensive amount of optimization methods and algorithms applied to the mission-planning layer. Some examples, and the problems on which they focus, are described next.

First, particle swarm optimization (PSO) is a population-based optimization algorithm which was proposed in 1995 by J. Kennedy and R. Eberhart [11]. More specifically, it is based on the flocking of birds. Most of the algorithm can be computed in a distributed manner, which is convenient in systems with limited computational power such as drone swarms. In terms of mission planning, this algorithm is used to solve the task assignment problem. For instance, Ref. [12] proposes a discrete PSO algorithm for assigning cooperating unmanned aerial vehicles to multiple tasks. In turn, Ref. [13] presents a centralized use of PSO for winged-UAV-swarm mission planning. Here, PSO is used as a quality mission start planner with high computational cost but little flexibility and resilience. Combination with other algorithms is also possible, as is the case in [14], where PSO is combined with genetic algorithms (GAs) by introducing crossover and mutation to the former. Similarly, in [15], simulated annealing (SA) is included in the inner loop of PSO to improve the local search and still preserve the fast convergence of PSO. Other possible improvements of the PSO algorithm are discussed in [16] (which uses the concept of Pareto equilibrium [17]) or in [18] (where sequential Monte Carlo is applied). The aim of the improved PSO proposed is to find the Pareto equilibrium for the multi-traveling salesman problem (mTSP), and it uses SMC to neglect low-efficiency particles and resample the higher-efficiency ones, improving the convergence rate of PSO in the latter optimization stages.

Another applicable optimization algorithm is the ant colony optimization (ACO), which is a swarm intelligence algorithm inspired by the behavior of ants. Ants achieve coordination through the segregation of pheromones, a chemical substance that can be perceived by other ants. In mission planning, it is usually applied to solve the TSP. Also based on biological principles (i.e., algorithms that follow some patterns found in nature, such as the conduct of some groups of living creatures which cooperate to achieve an objective), the wolf pack algorithm (WPA) adopts the bottom-up design principle and simulates wolves hunting cooperatively according to their division system of responsibilities. It it used to solve the task assignment problem. More specifically, it is very useful in military attacking missions where the target (one or several) is not static, as it represents a very similar case to the wolf pack trying to hunt a prey. However, it cannot be applied directly to a multi-task assignment problem, so it needs to be improved. For instance, in [19], PSO and a genetic algorithm were introduced into WPA to improve the convergence speed and solution accuracy.

Alternatively, simulated annealing (SA) (first introduced by S. Kirkpatrick in [20] as a meta-heuristic optimization approach inspired by the annealing of metals) has also been explored. For instance, Ref. [21] adds the cosine singularity clustering method (CSCM) as a task clustering algorithm before SA to ease the solving of the mTSP. Ref. [22] also uses a clustering algorithm, but in this case, it is k-means instead of CSCM. In [23], SA is used inside a GA to prevent the GA from stalling in local minima and, as a result, improves the compilation time with respect to a normal GA for the same task. Ref. [24] also uses SA to improve local minima blocking, but it uses the firefly algorithm instead of a GA. Ref. [15]

uses SA as the external part of a particle swarm optimization (PSO). Finally, Ref. [25] uses SA as a task assignment algorithm and compares the results with those obtained by the population-based algorithms genetic algorithm and differential evolution.

Finally, traditional linear programming (i.e., a well-known mathematical optimization method whose requirements are represented by linear relationships and the objective function is a linear function) and integer programming problems (i.e., a mathematical optimization in which some or all of the variables are restricted to be integer numbers) and mixed-integer linear programming (MILP) (i.e., both objective function and constraints are linear, and only a few of the variables are restricted to be integers) can also be applied. In terms of mission planning, MILP is commonly used in both task assignment and task scheduling. A representative example, which includes both steps, is presented in [26], where the problem is to find the optimal schedule to perform various tasks with different time windows at several locations making use of a fleet of fixed-winged heterogeneous UAVs.

Moving on from optimization methods, game theory is a mathematical discipline that studies the interaction (either competitive or collaborative) between two or more rational agents, commonly called players. Ref. [27] makes use of a mean-field game framework to reduce the communications between the UAVs of a massive swarm of drones, making its control possible. In [28], the authors model the distributed task assignment problem of UAVs as a coalition formation game, where they prove that the algorithm proposed can achieve the joint optimization of energy and task completion by the decision making of UAVs in finite iterations.

Artificial intelligence methods are gaining popularity for swarm mission planning. Particularly, reinforcement learning (RL) is an artificial intelligence field which consists of techniques where agents learn through trial and error in a dynamic environment (a more extensive discussion on RL will be conducted in Section 2.6). An example of an application of RL in a task assignment problem can be found in [29]. The authors model the problem in a 2D simulation, where the environment is represented as a grid and the drones are spawned at random locations within the environment. They see each task to be performed as a point of interest (PoI), uniformly distributed within the grid and with an assigned level of priority (low or high). Drones can move in eight possible directions: north, north-east, east, south-east, south, south-west, west, and north-west. In Figure 3, we show a representation of this environment which consists of a 5 × 5 grid, where stars represent PoIs and triangles are UAVs. Black stars represent low-priority-level PoIs, and red stars are high-priority-level PoIs. Stars are colored white once they are already mapped by a UAV. The reward function is defined in such a way that each UAV obtains a reward of $\alpha$ for mapping a low-priority point and a reward of $2\alpha$ for mapping a high-priority point, where $\alpha$ is a positive real number. In addition, if there is any high-priority PoI which is not mapped, the UAV receives a penalty, defined as the distance (norm $L_2$) between the UAV's location and the closest unmapped high-priority point. Finally, the agent receives a negative reward of value one at each time step, which encourages the policy to map the points faster to maximize the cumulative sum of rewards. The authors use different neural networks to compute the policy.

*2.3. Path-Planning Layer*

The next layer within the conceptual swarming architecture is path planning, which refers to mid-level planification, i.e., to the process of calculating a trajectory from an origin to an endpoint. These points might belong to a single task, whose execution may be implemented through a specific trajectory segment, or they may perform the travel between two different tasks. In the global scheme of swarm planning, path planning uses as its input the result of the preceding layer (mission planning), and its plan might be overridden by the lower real-time planning/control layers in charge of collision avoidance and formation control. Often, path-planning and mission-planning layers work collaboratively and in a tightly coupled architecture. There, the output trajectory is used in a feedback loop by the mission-planning layer to evaluate the costs and alternative task assignations and scheduling go through a path-planning calculation.

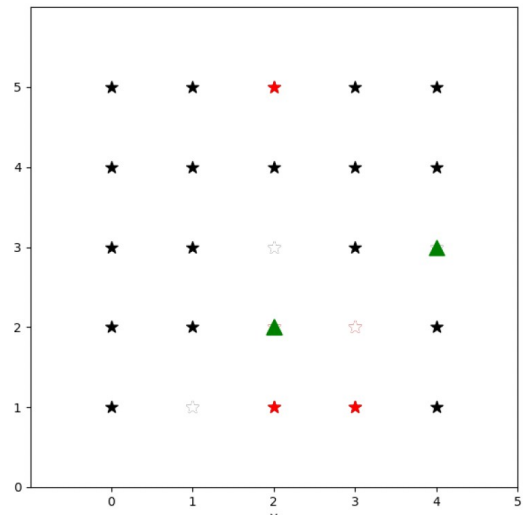

**Figure 3.** A 2D environment made up of a 5 × 5 grid (from [29]).

Path planning usually considers UAVs individually. This means that, even though all the UAVs of the swarm carry out this process, it is performed in parallel, with each UAV's path planning being independent. Because of this "individual agent approach", path planning is easier to distribute.

This planning step must consider some flight restrictions such as dynamic constraints, dependent on the type of UAV, or the potential existence of no-fly zones, due to threats, obstacles, changing terrain altitude, or civil security, among others. Also, it is possible that this layer performs some kind of strategic deconfliction between UAVs in the swarm, ensuring their flights very rarely lead to a loss in separation. As a result, the path planner should provide an efficient trajectory that meets all the constraints, reaches all task locations set by the mission planner, and includes all trajectory segments needed for the execution of each of the tasks.

Because of the existence of an extensive number of algorithms aimed at solving the path-planning problem, there are several classifications of them according to different criteria. Figure 4 shows the one that will be used in this section, along with reinforcement learning algorithms.

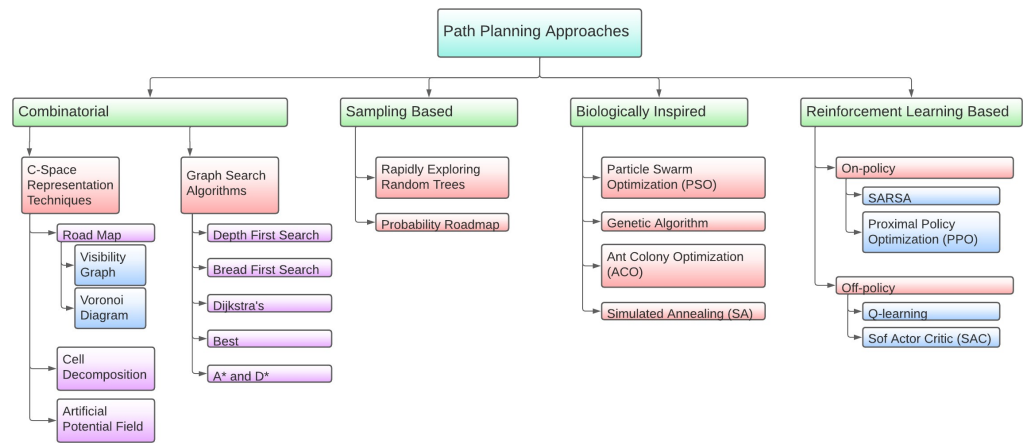

**Figure 4.** Classification of path-planning algorithms (inspired on [30]).

Starting with the first category, combinatorial path planning creates a route by resolving queries along the way. It can accomplish the required criteria without further adjustment in conventional algorithms. This category is subdivided into two different groups: C-space representation techniques and graph search algorithms.

Configuration space (C-space) provides detailed information about the position of all points in the system, and it is the space for all configurations. Therefore, the C-space denotes the actual free space zone for the movement of the robot and guarantees that the vehicle or robot does not collide with an obstacle. Road map, cell decomposition, and artificial potential field are examples of C-space techniques.

A road map is a map with a visual representation of paths used for automobile travel and aircraft navigation. It is a type of navigational map that commonly includes political boundaries and labels, making it also a type of political map.

In turn, potential field (PF) is an algorithm based on the attractive potential and repulsive potential generated in a certain space. Based on the potential field, the artificial potential field (APF) technique was developed for multi-robot systems. The idea is that the robot is a free particle susceptible to the forces of an "artificial field". However, the method presents some problems. Some environments, such as convex obstacles can create local minima in which the robot can become stuck. Therefore, a robust APF algorithm intended to be used in complex environments should be able to solve this problem. Ref. [31] solves the problem by introducing a variable force in the APF. In this way, the path-planning problem is transformed into an optimization problem, where the optimized variable is the variable force, and the cost function is the original APF. Also, Ref. [32] shows an example of a function that models this artificial field.

Finally, the cell decomposition (CD) representation mainly finds an obstacle-free cell and builds a finite graph for these cells. It consists in the discretization of the space into similar cubes (or squares in 2D cases) and their posterior evaluation according to a cost function (usually some kind of distance), assigning each cube a cost that will be used by the posterior path evaluation algorithm. Examples of cell decomposition can be found in [33], where octrees are applied with an APF as the path evaluation algorithm. Ref. [34] uses GeoSOT-3D, a space grid representation tool that can follow an octree hierarchy.

It is likely that one of the previous methods leads us to a graph, in which we need to find the best path between the set of possibilities this graph offers. This task is carried out by graph search algorithms, which refers to the process of visiting each vertex in a graph. The most common and used graph search algorithms are Dijkstra's algorithm, A* search algorithm, and dynamic programming.

Dijkstra's algorithm is an algorithm conceived by Edsger W. Dijkstra in 1956, aimed at finding the shortest paths between nodes in a graph. Dijkstra's original algorithm found the shortest path between two given nodes, but a more common variant fixes a single node as the "source" node and finds shortest paths from the source to all other nodes in the graph, producing a shortest-path tree.

The A* (or A star) search algorithm is a modification of Dijkstra's algorithm that considers the cost of reaching a determined node and the cost of reaching the final node from the examined node. In [35,36], the A* algorithm is used to find the optimal solution in the set of solutions obtained by a PRM. A simplification of the D* algorithm (which is an extension of the A* algorithm) is used in [37] for evaluating the set of solutions obtained by a modified PRM algorithm. The A* algorithm can consider different costs, for instance, in [38] it evaluates a Voronoi diagram to obtain a path that avoids determined danger zones and reaches a goal point. Moreover, Ref. [39] presents a version of the A* algorithm that takes into account the level of signal received to find a path that maintains 3G signal coverage in unfavorable conditions. Finally, in [40], some modifications to the A* algorithm are presented.

In terms of UAV path planning, dynamic programming refers to methods that evaluate a previous input graph by searching the least expensive path between different nodes, i.e., by reducing the problem of finding an optimum global path into finding various suboptimal paths. This is exactly how Dijkstra's algorithm works, so dynamic programming in UAV path planning follows a similar structure. Some variations are applied to the cost function used to evaluate the paths. In Dijkstra's algorithm, the metric used is the Euclidean distance, but the evaluation function can be changed to obtain paths that meet certain constraints. An example of the application of DP to the path-planning problem is shown in [41], where a swarm of drones is given the task of reaching several targets while avoiding threat zones, synchronizing minimum time arrivals on the target, and ensuring arrivals coming from different directions. In [42], DP is used to find the shortest path through a digital elevation map divided in 2D cells. The cost metric used is the distance. Ref. [43] presents a more complex case where path planning is carried out in a windy environment. To achieve the goal of finding the shortest-time travel path it includes the effect of the wind- and vehicle-related constraints in the cost function.

Moving to the next parent category in Figure 4, space sampling algorithms are those in charge of creating a feasible solution set for the path-planning problem. Probabilistic road maps (PRMs) are one of the most used types of algorithm in the creation of a solution set for the UAV swarm path-planning problem. A PRM is an algorithm that randomly generates routes that cover a series of goal points starting from different initial points. Therefore, a map of the environment is required, as well as the list of goal points and starting positions. In [36], a PRM algorithm is used in conjunction with the A* algorithm to find the optimized routes in a 2D environment. The same is performed in article [35] (by the same authors) but in a more complex 3D environment. Rapidly exploring random trees (RRT) is another extended space sampling algorithm. It was first introduced by LaValle in [44] as an alternative to PRM. The algorithm follows the same idea as PRM but, instead of creating random nodes and connecting them, it creates a growing tree of random nodes connected by free space edges until it reaches the goal. Therefore, there is a branch of the tree that is the path from the initial position to the goal. To separate the path from all the other branches of the tree, some path evaluation algorithm must be used. In [45], Dijkstra's algorithm is used for this purpose, but it requires a large amount of computational time, wasting one of the most attractive characteristics of RRT. To solve this issue, [46] proposes a faster pruning algorithm along with a path smoothing algorithm, aimed toward enabling the RRT path for winged UAVs, which are not able to cut sharp corners. Refs. [47,48] also take into account the physics and kinematics of winged UAVs and adjust RRT consequently. Moreover, Ref. [47] also adapts and tests it in dynamic environments (with pop up obstacles). Refs. [49,50] present modified RRT versions aimed towards shorter path lengths and algorithm calculation times. Ref. [51] proposes an RRT algorithm with a simple modification aimed towards consuming less energy, but it has not been extensively tested. A more complex RRT modification that uses Fisher information for target tracking is proposed in [52]. This version, as implied by its target tracking purpose, has dynamic capabilities. Finally, Ref. [53] modifies RRT to be used in uncertain dynamic environments.

As was the case in mission planning, biologically inspired algorithms may also be used. In this group, some of the algorithms mentioned in the previous section can be highlighted, such as particle swarm optimization, genetic algorithms, ant colony optimization and simulated annealing. Ref. [54] has been already mentioned as an application of ACO to mission planning, since it uses a genetic algorithm to perform the task scheduling, but it further represents a good generic example of ACO application to path planning. Here, the operational space is divided into cubes, whose corners are the possible waypoints for our goal. Depending on the environment, some of these cubes can be set as no-flight cubes (for instance, if they are buildings, geofences, etc.). Alternatively, Ref. [55] uses PSO to solve the problem of approaching three-dimensional route planning. In [56], a generic and simplified approach to using the GA for path planning is presented. A comparison between both previous methods (i.e., PSO and GA) is discussed in [57]. Moreover, it proposes a parallel implementation of both methods. Ref. [58], from the same authors, proposes a modified implementation of a GA for winged-UAV path planning that makes use of GPUs for a reduced processing time. It uses a GA to select the best path from a visibility graph. Other examples include [59], where they test a GA in an environment with obstacles with predictable movement. In [38], a GA is used with an APF, influencing the fitness function. The work proposes a GA and a multi-population GA for UAV path planning under emergency landing conditions. It also tests the system under different levels of wind. Ref. [60] proposes a GA called multi-frequency vibrational genetic algorithm that uses Voronoi diagrams. Moreover, it compares the results of the proposed algorithm with three additional GAs. As already mentioned, one of the greatest properties of the SA algorithm is its relative invulnerability against the local maximum problem, in contrast to some other algorithms such as genetic ones. This is the reason why the former is usually used as an aid in another optimization process, as in [61], which describes a path-planning process based on a genetic algorithm aided by simulated annealing.

Due to the complexity and computational cost of computing several paths for a swarm of UAVs, in recent years, the most common methods used in the resolution of this problem are those based on artificial intelligence (AI). All RL algorithms applied to the path-planning problem follow a common structure. The features that make the difference are learning strategies, which follow different policies that make it possible to deal with different kinds of problems. The two most common policies in UAV path planning are Q-learning and SARSA (which will be described in Section 2.6). For instance, in [62], a Q-learning algorithm is proposed to solve the path-planning problem for UAVs in an unknown antagonistic environment, where the aim of the agent is to obtain the maximum cumulative rewards in the interaction with this environment. The authors begin by defining the environment as a limited 2D space with one target and three threat sources. A Q-value update is made according to Equation (1). Ref. [63] also uses Q-learning to calculate UAV paths under threats. Ref. [64] extends [62] to dynamic environments. In [65], a double-duelling deep Q-network is used for path planning. Alternatively, in [66], UAVs receive information from a certain number of routers placed on the ground and the authors use SARSA to teach the swarm how to maximize the network lifetime. Moreover, Ref. [67] uses SARSA as a method for path planning and obstacle avoidance in a dynamic environment.

Focusing on surveillance use cases, in [68], the authors address a problem similar to ours, a multi-target tracking problem using a swarm of UAVs which fly at a fixed altitude. Nevertheless, the authors assume a constant speed for the UAVs, while for us the control of the speed is one of the actions provided as an output of our models. Furthermore, they do not take into account obstacle avoidance, which is a critical point in our solution. Meanwhile, Ref. [69] aims to design an autonomous search system for a swarm to localize a radio-frequency mobile target. The authors consider a search-and-rescue application scenario in which UAVs equipped with omnidirectional received-signal-strength sensors have to navigate towards a single ground target which radiates radio-frequency signals. In our case, we extend the problem to also keeping coverage of the targets once they are first detected, while avoiding collisions with hazards. In [70], the authors focus on the problem of cooperative multi-UAV observation of multiple moving targets. They optimize the average observation rate of the discovered targets and also emphasize the fairness of the observation of the discovered targets and the continuous exploration of the undiscovered targets. For this purpose, in the article referred to the authors work with a discretized two-dimensional rectangle divided into cells through which four observation-history maps are generated and merged. Actions also rely on these cells, whereas our system considers a much lower-level definition of actions, again enabling the joint solution of the problems at the different swarming levels.

### 2.4. Formation-Control/Collision-Avoidance Layer

During a swarm mission, following previously calculated UAVs' trajectories with some errors, UAVs may lose separation either with other UAVs in the swarm or with obstacles. To avoid this loss of separation, in addition to the aforementioned strategic deconfliction approaches in the planning phases, there are typically real-time tactical conflict detection and resolution system, which enable avoiding collisions both with neighbors and some obstacles. These obstacles can be classified as static (ones whose location is predefined before the mission starts, such as buildings, terrain, etc.) and dynamic (ones that appear in the course of the mission, such as threats or possible collisions with other UAVs). The dynamic nature of some obstacles is another reason why a real-time control layer is needed in the planning process of a swarm mission. Therefore, this layer is aimed at avoiding (if possible, in an autonomous way) collisions and conflicts both between UAVs themselves and between the environment and UAVs.

An example of a formation-control problem is shown in [71], where a method of multi-UAV cluster control, based on improved artificial potential field, is proposed. As for RL-based methods, in [72] a deep Q-learning algorithm is proposed to solve the problem of avoiding collisions between UAVs whose aim is to reach a goal position. We can see another example of the application of Q-learning in [73], where a Q-learning algorithm is used for both path planning and collision avoidance at the same time. In [67], the authors also show how to apply a deep SARSA algorithm to solve the dynamic obstacle avoidance problem.

## 2.5. Swarming Survey Summary and Conclusions

As shown in the previous sections, in the literature there exists an extensive amount of optimization methods and algorithms applicable to the different swarming control layers. Table 1 summarizes which kind of algorithms can be applied to each layer.

**Table 1.** Methods used for different swarm control layers.

| Method | Mission Planning | Path Planning | Formation Control/Collision Avoidance |
|---|---|---|---|
| Reinforcement Learning (RL) | ✓ [29] | ✓ [62–70] | ✓ [67,72,73] |
| Particle Swarm Optimization (PSO) | ✓ [11,13–18] | ✓ [55,57] | – |
| Ant Colony Optimization (ACO) | ✓ [54] | ✓ [54] | – |
| Wolf Pack Algorithm (WPA) | ✓ [19] | – | – |
| Genetic Algorithm (GA) | ✓ [14,19,23,54] | ✓ [38,56–60] | – |
| Simulated Annealing (SA) | ✓ [15,21,23–25] | ✓ [61] | – |
| Game Theory | ✓ [27,28] | – | – |
| Dynamic Programming | – | ✓ [41–43] | – |
| Mixed-Integer Linear Programming (MILP) | ✓ [26] | – | – |
| Artificial Potential Field (APF) | – | ✓ [31,32,36,38] | ✓ [71] |
| Dijkstra's Algorithm | – | ✓ [45] | – |
| A* Search Algorithm | – | ✓ [35–40] | – |
| Probabilistic Road Map | – | ✓ [35–37] | – |
| Rapidly Exploring Random Trees | – | ✓ [44,46–53] | – |
| Octrees | – | ✓ [33,34] | – |

It is possible to perform every step in the planning process of a swarm mission using classic algorithms (i.e., optimization algorithms, graph-based methods, etc.). However, these methods have some limitations, often related to complexity or computational cost. This is typically the case when dealing with multi-UAV missions, since there exist computationally constrained agents which perform a certain number of tasks, often in real time.

Reinforcement learning (RL) algorithms are suitable for this kind of problem, especially when it comes to dynamic environments. The definition of RL itself enables the adaptability of this method to new circumstances. Furthermore, the capability of combining RL with deep learning (deep RL) makes it even more appropriate, since neural networks greatly reduce the computational cost of the algorithms and accelerate the convergence at the same time. This is the reason why RL and deep RL methods are particularly relevant [74]. These methods will be discussed in detail in the next subsection.

In contrast with related works in the literature, the system proposed in this paper uses deep reinforcement learning algorithms to obtain behavior models able to solve the whole surveillance problem, taking into account obstacles of the environment as well as more realistic modeling of agents and target dynamics, sensor coverage, etc.

*2.6. Reinforcement-Learning-Based Methods*

Finishing the review of the current state of the art, reinforcement learning (RL) will be discussed, as it is the main AI method used in the proposed swarming architecture. The goal of RL is to optimize the behavior of an agent according to the evaluative feedback received from the environment, i.e., to find an optimum policy for achieving a certain goal. Therefore, one vital part of RL is the evaluation policy (or learning policy), which is the way the response from the environment is evaluated and how the agent's performance is adjusted according to the reward. The learning policies used in RL can change, but the general structure is similar in all RL techniques.

RL is a semi-supervised technique. The main difference between RL and supervised learning is that in RL, the agent does not know about the correct action to be performed for a given input. Instead, it tries to learn the optimal action based on the reward it receives.

Problems solved by RL algorithms are very similar to the Markov decision process (MDP), involving the following information:

- Finite set of states $S$.
- Finite set of actions $A$.
- Transition function $P_a(s, s^t)$, defined as the probability that performing action $a$ in state $s$ leads to state $s^t$ in the next time step.
- Reward $R_a(s, s^t)$, provided as a result from moving from state $s$ to state $s^t$ due to action $a$.

Therefore, being in the current state, the agent must select an action following a policy (which represents the agent's behavior). This selection gives us a reward (provided by the reward function) and leads to another state (following the state transition probability) according to the environment dynamics.

RL algorithms can be divided into two categories: on-policy learning and off-policy learning. The first category learns only from data gathered in the current observation and it is more conservative, while the latter learns from the information obtained in all the previous steps and it is greedy in its decisions, which means that it will assume that the decision taken is the one with the highest reward.

The most common algorithms in RL are Q-learning (categorized as off-policy learning) and state–action–reward–state–action (classified as on-policy learning).

Q-learning [63] follows a model-free strategy, so it updates its knowledge following a policy purely of trial and error. It is an off-policy algorithm, which means that data generation, and therefore, policy updating, is determined by a different policy from the one used for behavior. This allows the system to use data gathered from previously used policies.

It is based on the filling of a Q-table, which contains the values of each pair of state–action, i.e., for each state (rows) it summarizes the value of taking each possible action (columns). These values are known as Q-values, which are updated following a Q-function defined by the equation

$$Q(s, a) = Q(s, a) + \alpha \left[ r + \gamma \max_{a^t} Q(s^t, a^t) - Q(s, a) \right] \quad (1)$$

where $s$ is the current state, $a$ is the current action, $s^t$ is the next state, $a^t$ is the next action, $r$ is the immediate reward, $\alpha$ is the learning rate, and $\gamma$ is the discount factor ($\gamma < 1$). The Q-learning target policy is always greedy, since to predict the Q-values it will always assume that the action taken is the one with the highest quality, i.e., it always takes the maximum between all $Q(s^t, a^t)$ possibilities. We show the pseudocode of the Q-learning algorithm in Algorithm 1.

---
**Algorithm 1** Q-learning

　　initialize $Q(s, a)$ arbitrarily, where $s$ denotes the state of the agent and $a$ denotes the action
　　**for** each episode **do**
　　　　initialize $s$
　　　　**while** $s$ is not terminal state **and** steps number < max steps number **do**
　　　　　　choose $a$ from $s$ using policy derived from $Q$
　　　　　　take action $a$, observe reward $r$, and next state $s^t$
　　　　　　$Q(s, a) \leftarrow Q(s, a) + \alpha[r + \gamma \max_{a^t} Q(s^t, a^t) - Q(s, a)]$
　　　　　　$s \leftarrow s^t$
　　　　**end while**
　　**end for**

---

On the other hand, state–action–reward–state–action (SARSA) [67] is an on-policy RL algorithm. This means, contrary to Q-learning, it has a unique policy for data generation (and, consequently, policy updating) and for behavior. Hence, the system can only learn from the data gathered by the current policy and new updated policies need to gather their own data. SARSA, as well as Q-learning, uses a Q-table with a Q-value for each pair of state–action. Instead, SARSA's target policy is the same as its behavioral policy, which means that to predict the Q-values it will assume that the action taken is coherent with the actual policy. In terms of the Q-function, this leads to the following equation:

$$Q(s, a) = Q(s, a) + \alpha \left[ r + \gamma Q(s^t, a^t) - Q(s, a) \right]. \quad (2)$$

where, as in Equation (1), $s$ is the current state, $a$ is the current action, $s^t$ is the next state, $a^t$ is the next action, $r$ is the immediate reward, $\alpha$ is the learning rate, and $\gamma$ is the discount factor ($\gamma < 1$). Notice that the difference with Q-learning is the maximum taken in the term multiplied by the discount factor $\gamma$. We show the pseudocode of the SARSA algorithm in Algorithm 2.

---
**Algorithm 2** SARSA

　　initialize $Q(s, a)$ arbitrarily, where $s$ denotes the state of the agent and $a$ denotes the action
　　**for** each episode **do**
　　　　initialize $s$
　　　　choose $a$ from $s$ using policy derived from $Q$
　　　　**while** $s$ is not terminal state **and** steps number < max steps number **do**
　　　　　　take action $a$, observe reward $r$, and next state $s^t$
　　　　　　choose the next action $a^t$ from $s^t$ using policy derived from $Q$
　　　　　　$Q(s, a) \leftarrow Q(s, a) + \alpha[r + \gamma Q(s^t, a^t) - Q(s, a)]$
　　　　　　$s \leftarrow s^t$
　　　　**end while**
　　**end for**
---

In addition to Q-learning and SARSA, many more algorithms have been developed in order to improve these ones or achieve different features. For instance, proximal policy optimization (PPO) [75] and soft actor–critic (SAC) [76] are two different RL algorithms, classified as on-policy and off-policy, respectively. PPO is a policy gradient method that alternates between sampling data from the agent–environment interactions and optimizing a "surrogate" objective function (*L*) through stochastic gradient ascent in a series of minibatch updates, as explained in Algorithm 3. The *L* function is a version of the policy update function (i.e., it pushes the agent to take actions that lead to higher rewards), but constrained so that large changes are avoided in each step.

---

**Algorithm 3** PPO
---
  **for** each iteration **do**
    **for** *T* time steps **do**
      Run previous policy $\pi_{\theta_{old}}$ and retrieve (state, action, reward)
    **end for**
    Compute advantage estimates $\hat{A}_1,\ldots,\hat{A}_T$ of each taken action
    Optimize *L* function with regard to $\theta$ in *K* epochs and minibatch size < *T*
    Update policy parameters: $\theta_{old} \leftarrow \theta$
  **end for**

---

Before proceeding with the deep reinforcement learning (deep RL) definition, we need to provide a short introduction to deep learning (DL), which is a class of machine learning (ML) algorithms that use multiple layers of neural networks to progressively extract higher-level features from the raw input. We obtain deep RL methods when deep neural networks are used to approximate any of the features mentioned before: policies, state transition function, reward function, etc. In terms of Q-learning and SARSA, deep neural networks can even be applied to replace the estimation of Q-values or to avoid the storage of them in a Q-table.

Deep RL emerges because of the unfeasible computational cost of applying RL by itself in complex environments. As neural networks achieve good approximations for nonlinear functions, they can be used as a state evaluation function and policy function with reduced computational costs. Figure 5 shows the structure of an RL model that estimates the policy function with a deep neural network.

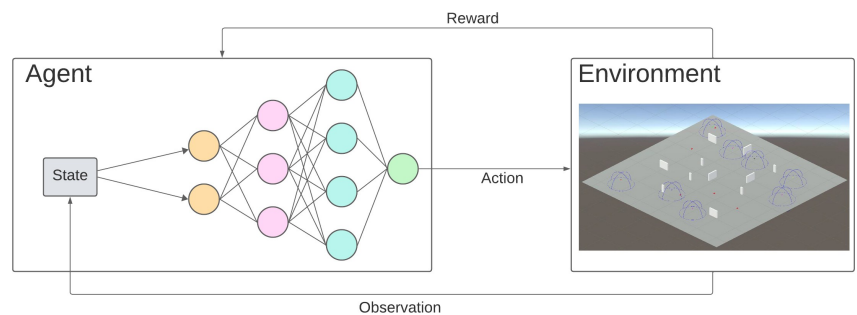

**Figure 5.** Structure of deep RL with a policy neural network.

## 3. Proposed System

In this paper, the focus is on the application of the aforementioned techniques to demonstrate the self-organization capacity of a UAV swarm to carry out ground surveillance, tracking, and target designation missions in a designated area. The following methodology was followed to develop the proposed swarming system:

1. First, a representative use case of the surveillance problem was compiled, describing the expected concept of operation of the swarm, the types of available drones, the targets of interest, limitations, etc. As a result, a set of requirements for the swarm was derived, as discussed in Section 3.1.

2. Considering the problem description, a system-level architecture was designed, as described in Section 3.2. The architecture adapts the conceptual architecture previously depicted in Figure 1 to a centralized swarm organization. Considered tasks, information exchange between the swarm elements, and selected algorithms for each element are also discussed here.
3. Then, the behaviors implemented with DRL were trained using a simulation environment. A novel hybrid methodology was followed, consisting in dividing the drone functionality into individually trainable models. Then, for each model, an iterative training approach was adopted consisting of increasingly adjusting the scenario complexity and rewards until the desired behavior is obtained. This process is explained in Section 3.3.
4. Finally, the implementation was validated using a set of simulation scenarios. For this, we propose relevant performance indicators for the considered problem. These metrics and results are discussed in Section 4.

*3.1. Surveillance Application Description and Requirements*

The surveillance mission is specified by defining the spatial volume (polygonal plus height limits) in which the surveillance task must be carried out. The targets will move on the ground, just below our flight volume, in the so-called operation area. The swarm will be made up of several drones and a central swarm controller (SC) placed on the main platform that deploys the swarm. So, in our case we will follow a centralized swarm organization.

The UAV swarm mission may be understood as being composed of the following steps, comprising a collection of tasks:

1. Deployment of the swarm: the swarm of drones is released from a platform.
2. Target search and acquisition, where swarm assets, holding suitable sensors for exploration, search for the targets that are moving in the area of interest.
3. Target tracking, after target acquisition, where the SC assigns to some of the UAVs the tasks to follow and track the detected targets while the swarm maintains the ability to acquire new targets.
4. During the continuous tracking of targets, both the SC and the UAVs' control systems shall take into account the possibility of losing the target. Loss of target sight can happen due to occlusion, which would occur in low-altitude ground environments where obstacles are present, to evasive maneuvers by the target, or by errors in the swarm control functionality. Swarm control must reassign tasks to reacquire the target. Tracking UAVs may also be reassigned to either tracking or search as the target leaves the area of interest.
5. Once the time is over, there shall be a retreat phase of the UAV swarm to the pick-up point. The swarm's intelligent control shall manage the path planning for such a retreat sequence if the follow-up mission is interrupted. It then shall adapt to the current swarm status and reassess the UAVs' trajectories towards a safe point.

It should be noted that at a given time, some UAVs will be at different states (related to the 2nd, 3rd, and 4th states in the previous list), so after deployment a given UAV can be either performing search tasks, tracking tasks, or reacquiring the target.

For this surveillance scenario, the UAVs to be considered are small vehicles (<100 kg), capable of hosting sensors with sufficient range to cover an area of interest with a span of several kilometers. Also, platforms with a several hours of flight autonomy will be chosen, and therefore, fixed-wing UAVs powered by internal combustion engines will be selected. Different scenarios could lead to different UAV selections and will demand a different RL training process, although the same principles and approaches to system design will hold. The use of fixed-wing unmanned aircraft imposes restrictions on the maneuvers available during operation. Slow speeds are problematic, and continuous motion is needed in contrast with rotary wing platforms. The swarm path-planning control shall take into consideration these motion restrictions to fulfill mission tasks and establish collision-avoidance strategies. UAVs can move anywhere within the volume of interest.

The surveillance scenario will be defined as a rectangular area of interest, with the following conditions (with a direct impact in training and evaluation of the swarming solution):

- The paths the ground targets may follow are not constrained. The trajectories of the targets are combinations of straight sections joined by curves in which the direction is changed. Speeds and accelerations will be typical for the interest ground targets (vehicles, individuals, etc.).
- The scenario might contain sparse buildings that could occlude the ground targets.
- The meteorological conditions in the scenario are assumed to be benign, without adverse weather conditions such as heavy rain or fog, and the system is oriented to daily surveillance. We can, therefore, assume camera sensors will allow correct target detection if the targets are within range of the sensor.
- UAVs can carry two complementary types of sensors for target detection: short-range millimeter radar and optical cameras. The camera can measure the target's direction and the radar target's distance, which enable a 3D location of the target's position.
- For obstacle avoidance, an additional proximity sensor is implemented in each UAV.
- The altitude at which the UAVs fly is the same for all of them and it is not allowed to be modified.

A target, although within sensors' coverage range, may be hidden behind obstacles (i.e., buildings). In these cases, the target is not detectable, and tracking can be interrupted. For this reason, within the demonstrator it is necessary to include these phenomena to test if the swarm control system can recover the tracking (reassigning UAV, reorienting the sensors, or monitoring the areas bordering the shadow zones). In our mission, ground vehicles are considered as targets. The process to detect/classify a target is a sensor design problem, which is not the focus of this study, where perfect sensors will be assumed.

Targets will enter the area of interest at a certain rate. In this way, in the scenario, there will be targets being tracked while new ones appear. The SC will need to reassign the tasks of the swarm UAVs to incorporate the new targets into the tracking mission. If the resources are not enough (the maximum number of drones has been reached), the assets tasks are reassessed and the complete mission will not be covered with the same quality.

In addition, it is assumed that the UAVs are equipped with a navigation system that allows them to be located, relative to the area of interest, with a negligible error. Again, the reason is that the scenario aims to demonstrate the use of RL algorithms in swarm control, and navigation error is assumed to be a second-order problem for this.

*3.2. System Architecture*

To solve the surveillance problem described in the previous section, a proof-of-concept system has been designed which will showcase the use of reinforcement learning to solve the swarming problem at different planning layers. As already discussed, swarming can be addressed at various abstraction levels. One approach might choose to directly coordinate drones at the flight-execution level by providing flight commands to each individual UAV so that the overall objective is achieved. Here, we followed the centralized-layer architecture presented in Section 2, with a higher-level coordination (SC), where the reconnaissance and tracking mission is decomposed in a set of well-defined tasks to be assigned to a UAV. Then, each UAV is responsible for fulfilling the allocated task by dynamically generating the required flight control instructions. In fact, in our solution we will assume the tasks are provided to the drones one by one, and each drone does not, therefore, have the possibility to schedule its execution by permuting tasks' execution orders. So, the central SC performs the mission-planning layer function, while the UAVs' agents are a slave of the SC, implementing the following task execution and covering both the path-planning and collision-avoidance layer functionalities.

Following this division, a two-layer architecture (depicted in Figure 6, and further refined in Figure 7) is designed to implement the swarming system. It consists of the following:

- Swarm controller (upper layer): This is the centralized entity coordinating the UAV swarm. It periodically receives information from each UAV agent regarding its state and its surveillance information. Tasks are generated and allocated to drones as required, so that the surveillance and tracking objectives are fulfilled. Those tasks are communicated to the lower layer so that they are executed by the drones. In our proof of concept, this functionality will be rule-based and deterministic, although in future iterations we would like to extend the use of RL also to this layer.

- UAV agents/entities (lower layer): A set of autonomous and independent drones conforming the UAV swarm. Individual UAV behavior is dependent on the task that has been assigned by the controller. That is, drones dynamically adapt their trajectories to fulfill the task. As the task is executed, a UAV retrieves operational information (e.g., target positions) with simulated onboard sensors and forwards it to the swarm controller. In our proof of concept, this is the functionality to be implemented using RL.

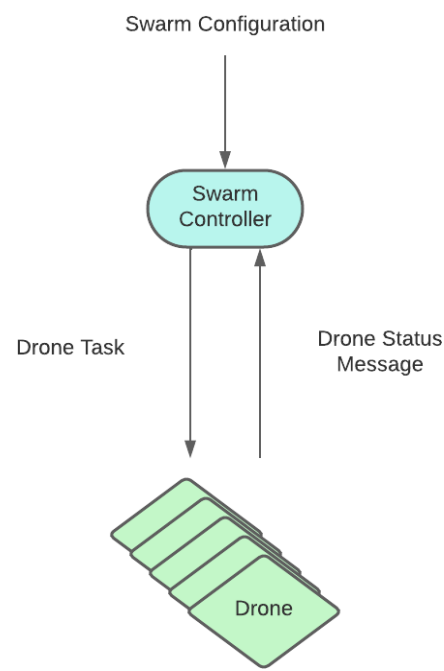

**Figure 6.** UAV swarm system. Coordination architecture.

In the system, and following the model discussed previously in this section, tasks are the atomic assignment units into which the whole swarm mission can be decomposed. The complex high-level swarm mission (in our case a surveillance mission) is divided into simpler tasks achievable by a single UAV:

- Search task: The overall swarm search function aims to achieve a complete level of recognition of the search area, covering it as fast as possible. From the UAV perspective, a search task consists of two processes that must be carried out in an orderly way:
    1. An initial movement to the assigned search area, which is a reduced section of the complete mission search area (if it is out of its assigned search area when it is assigned this task).
    2. The subsequent reconnaissance and target search phase, within the assigned search area itself, using search patterns.

For this purpose, the assigned UAV shall follow a criterion that guarantees the total exploration of its assigned search area periodically, or that reduces to the minimum time between revisits for each location within the assigned search area.

During the search the UAV can detect targets. When a UAV finds a target, it will notify the controller about the position, velocity vector, and identification/classification of the target. On its end, and as a result, the swarm controller will generate new tasks (namely, tracking tasks) to ensure that identified targets are followed.
- Tracking task: The main objective of the tracking task is to ensure the tracking of the targets. The tracking itself will consist of keeping the targets within the field of view of the UAV assigned to continuously track them. For each UAV, this task consists of two processes that must be carried out in an orderly way:
    1. An initial movement to the assigned target's last known position (if the target is not within the sensor coverage of the UAV).
    2. The subsequent tracking phase itself.

    Within the tracking phase, it may happen that the target is hidden behind an obstacle in such a way that the UAV loses sight of it. This may result in the UAV having to search for the target again while the tracking task is in process. This target re-acquisition search, integrated into the tracking tasks, is not similar to the one performed in the search task, as the latter pursues the goal of inspecting an entire area in the shortest possible time. In this case, the UAV must search around the last point where the target was detected, since it is known that the UAV is close to that point. In the initial movement of the UAV towards the assigned target, a similar local search may also need to be performed around the last available detected position of the target.

The swarm controller is in charge of generating and allocating search-and-tracking tasks through a dynamic process, so that the whole swarm performs the surveillance mission. Apart from the task allocation subsystem, additional logic is required in the swarm controller to interface with the lower layer. Periodic information from messages from drones is to be received and processed in internal data structures. These data structures form a fused operational picture with all the information received from the swarm and serves as an input for the SC algorithm. Once tasks are allocated, additional interfacing logic is also needed to forward assigned tasks to each UAV. This fused operational picture is composed of the following information:

- Status of each of the drones of the swarm. UAVs periodically send a UAV status message containing:
    - UAV position and velocity.
    - Current task being performed by the UAV.
    - Current list of detections: targets that are in the field of view of the UAV.
- Current visible targets: A list of the targets detected by the whole swarm. For each target, it contains a fused position estimation from all its detections by the swarm.
- Current search areas: This consists of the set of sub-areas the swarm operational area is divided into for searching task purposes.
- Current task assignations: As tasks are created and assigned, the swarm controller keeps a record of the last assignments provided.

*3.3. Implementation and Training Methodology*

The current implementation of the SC consists of a series of well-defined rules that generate new tasks and assignments following a series of objectives, priorities, and restrictions:

1. The whole operational area must always be covered by a set of search tasks.
2. As new targets are discovered, the swarm shall start tracking them while redistributing the search areas (generating new ones if required) between the remaining non-tracking drones.
3. It is assumed that each UAV can only fulfill one task at a time. Target tracking has more priority than search tasks. Also, we assume one UAV can track, at most, one target. Therefore, once a UAV is assigned to tracking a specific target, no other tasks will be assigned to that agent (UAV) unless tracking is unsuccessful.
4. It is also assumed that the number of drones is greater than the number of targets.

In turn, UAV agents need to carry out the tasks specified by the swarm controller. Therefore, they must be able to conduct target searches covering specified areas and track the movement of a target assigned to them. This behavior is achieved using deep reinforcement learning. One possible approach, typically used in the literature, would be to train a single model that considers all tasks. However, this technique requires complex training scenarios that must consider comprehensive use cases for all relevant tasks. Alternatively, we propose the usage of hybrid AI to combine the intelligence of different targeted models for each of the possible tasks. In this approach, the overall agent behavior is divided into different atomic sub-behaviors, named sub-agents. Each sub-agent corresponds to a model tuned to fulfill a given task, including specific observation and action spaces that are relevant for that task. Then, those sub-agents are associated to form the global behavior. Different rules and heuristics can be used to choose a single sub-agent or to combine the output of different sub-agents to guide the agent's behavior.

Three critical aspects, among others, are involved in the training process of an RL model (and particularly of sub-agents): observations, actions, and rewards. Observations are the inputs of the policy, and they need to be selected carefully since they form the basis on which the actions to be taken are decided. An action is an instruction from the policy that the agent carries out. Actions are the output policy and they will directly affect the flight control of the UAV. Once an action is taken, the sub-agent needs to know whether or not it was right in its decision. Hence, a positive or negative reward is given depending on the result replicated by this action on the environment. For this reason, rewards are also critical to assess the performance of the policy, and, therefore, to drive the training process. The optimal configuration of these three aspects might be different for each considered task of the model. In fact, they must be closely related and tuned to the considered tasks, and its correct design constitutes one of the main challenges when implementing RL models. This burden can be reduced by our proposed approach, as it enables decoupling the training for different tasks. In our design, each sub-agent may be designed and trained independently, as they are focused on different exclusive parts of the flight/UAV mission. Thus, specific purpose-built training scenarios can be used for each sub-agent. Moreover, this approach can also provide partial explainability of the agent's behavior, which is relevant in the aviation domain.

Coming back to the use case, each agent includes three sub-agents: one for search, another one for tracking, and a third one for obstacle avoidance. The sub-agents deployment into the general swarming process layers is depicted in Figure 7. These sub-agents are activated or deactivated depending on the requirements of the task associated with the agent and the environment conditions (only one sub-agent is activated at any given time). That is, if the swarm controller assigns a search task to the agent, then the search sub-agent is activated and it will manage the actions taken by the UAV, and the same will happen with the tracking task sub-agent. The obstacle avoidance sub-agent acts slightly different, since it is activated when the proximity sensor detects an obstacle or another agent, and resumes control of the task execution once the obstacle is avoided. Next, a description on how these sub-agents manage the UAV movement is provided.

Each of the three models associated with the sub-agents implement different policies that require different inputs (observations) and may also produce different outputs, corresponding to actions that control the flight of the agent. To approximate the ideal function that maps a sub-agent's observations to the best action, an agent can take in a given state (its policy); a neural network is used, following the same kind of deep RL structure introduced in Figure 5. As for the choice of this network, there exist many different possibilities depending on the properties of the environment and observations. In our specific use case, and in view of the simplicity of our observations, an artificial neural network (ANN) is selected. Currently, this ANN is made up of an input layer (observations layer), two fully connected layers (hidden layers), each of which is composed of 128 neurons, and an output layer.

To provide the ANN network with the required training data, it is necessary to simulate an interactive environment, where an agent can learn using feedback from its own actions and experiences. To achieve this, Unity [77] was used. This is a 3D graphical engine able to create different functionalities, physics, and scenarios. This platform includes a Unity machine learning toolkit (ML-Agents), an open-source project designed to create intelligent agents in simulated environments. This toolkit includes some Python trainers (based on PyTorch) with machine learning algorithms enabling agents' training. Our proof of concept used the default training algorithm, which is the well-known proximal policy optimization (PPO) (already described in Section 2.6), since it strikes a balance between ease of implementation, sample complexity, and ease of tuning [78].

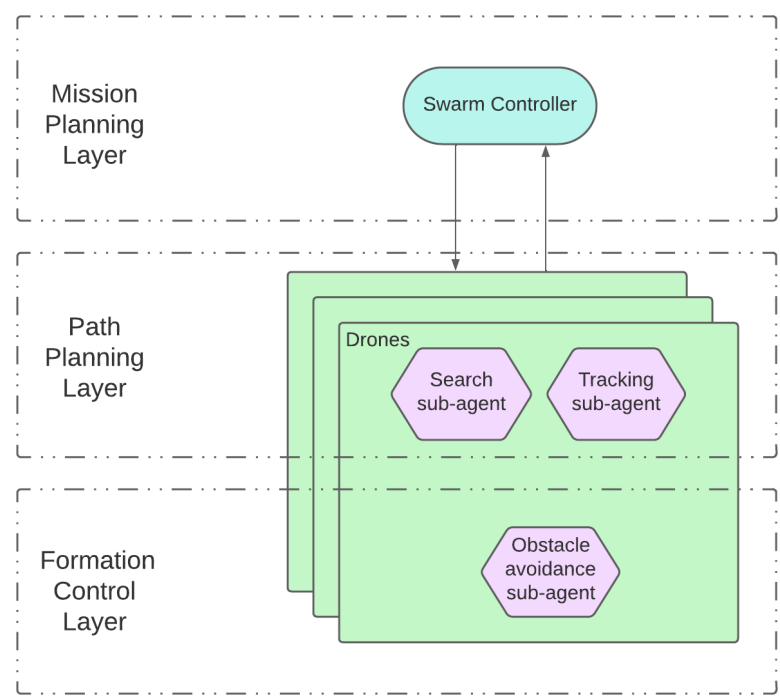

**Figure 7.** Deployment of the sub-agents in the different layers.

In the next sub-sections the training of the three sub-agents will be described, covering the associated observations, actions, and rewards. The predefined altitude assumption laid out previously implies that obstacles and other UAVs need to be avoided horizontally, and it also results in a much increased UAV conflict rate with respect to an alternative problem where different altitudes would be possible, or where 3D maneuvering could be used. On the other hand, this simplifies the actions and observations, as every observation and action taken by the sub-agents that involves a vector or a position can be considered as being two-dimensional. A cinematic model is used to adjust the agents' position based on the network-provided control parameters, that are different for each agent.

3.3.1. Search Sub-Agent Training

The search sub-agent implements the behavior of a UAV with a search task assigned. Thus, it is expected that the UAV flies to the assigned search area and explores it, periodically implementing different search patterns.

At a lower level, our model sets a rectangular grid on the search area, dividing it into small square cells, with the sizes related to the sensor coverage radius. The ultimate goal during training is to learn how to explore every cell in the designated search area in the task definition at least once within an established time limit. The observations, actions, and rewards defined within this aim are described below.

- Observations.
    1. Agent position.
    2. Agent velocity vector.
    3. Directional vector of the complete operation area center with respect to the UAV position.
    4. Operation area size.
    5. Directional vector of the task search area center with respect to the UAV position.
    6. Search area size.
- Actions.
    1. Heading control: a 2D vector indicating the direction to be followed by the UAV.
    2. Velocity control: the speed will be controlled indirectly through the thrust of the UAV by applying a force to the UAV, resulting in either acceleration or deceleration. The range of speeds achieved by the UAV in this way is constrained within 35 to 120 m/s.
- Rewards. Table 2 summarizes the rewards given to the search sub-agent during the training stage so that the expected search behavior is reached.

**Table 2.** Training rewards used for the search sub-agent.

| Trigger | Reward | Explanation |
| --- | --- | --- |
| Pass at least once through each cell in the task search area within the time limit. | +1 | Aim: to reach the task search area, stay inside, and explore it. |
| Entering the task search area. | +0.1 | |
| Leaving the task search area. | −0.25 | |
| Reach max step. | −0.5 | Aim: to avoid stand-by or loitering positions. |
| Entering the operation area. | +0.1 | Aim: to stay inside the operation area. |
| Leaving the operation area. | −0.25 | |

3.3.2. Tracking Sub-Agent Training

The tracking sub-agent is expected to make the UAV reach the target's last known position and track the target (i.e., follow it) as it moves. Additional behaviors such as initial target finding and lost target local search are also expected. During this part of the training process, targets initially move at a constant speed in straight lines between random points spread over the operation area. The ultimate goal of the tracking training is to learn to reach the target, and then keep it in the UAV field of view without interruption during an established (large) number of steps. The observations, actions, and rewards defined within this aim are described below.

- Observations.
    1. Agent position.
    2. Agent velocity vector.
    3. Directional vector of the complete operational area center with respect to the agent position.
    4. Operational area size.
    5. Directional vector of the target's last known position with respect to the agent position.
    6. Target's last known velocity vector.
    7. A Boolean indicating whether or not the target is being detected.
- Actions.
    1. Heading control: same meaning as in the case of search agent.
    2. Velocity control: same meaning as in the case of search agent.

- Rewards. Table 3 summarizes the rewards given to the tracking sub-agent during the training stage so that the expected tracking behavior is reached.

**Table 3.** Training rewards used for the tracking sub-agent.

| Trigger | Reward | Explanation |
| --- | --- | --- |
| Reach a certain number of consecutive steps detecting the task target. | +1 | Aim: to reach the search area, stay inside, and explore it. |
| The task target enters the sensors' coverage. | +0.1 | |
| The task target leaves the sensors' coverage. | −0.25 | |
| Reach max step. | −0.5 | Aim: to avoid stand-by or loitering positions. |
| Entering the operation area. | +0.1 | Aim: to stay inside the operation area. |
| Leaving the operation area. | −0.25 | |

3.3.3. Obstacle Avoidance Sub-Agent Training

The obstacle avoidance sub-agent is expected to take action when an obstacle or another agent is detected by the proximity sensor. To make an agent learn this behavior, a target position is placed randomly in the operation area and the UAV is forced to reach it moving in a straight line. If an obstacle or another agent is detected, the obstacle avoidance sub-agent takes control of the movement until the hazard is completely avoided, and the UAV continues its displacement towards the target point. The observations, actions, and rewards defined to this aim are described below.

- Observations.
    1. Agent position.
    2. Agent velocity vector.
    3. Directional vector of the target position with respect to the agent position.
    4. Proximity sensor detections.
- Actions.
    1. Heading control: three discrete actions to control the yaw can be taken. The value 0 is used when a forward movement is wanted, so the UAV follows a straight trajectory with the yaw equal to 0. The values 1 and -1 are taken when a rotation in the yaw is wanted.
    2. Velocity control: same meaning as in the case of the search agent.
- Rewards. Table 4 summarizes the rewards given to the obstacle avoidance sub-agent during the training stage so that the expected behavior is reached.

**Table 4.** Training rewards used for the obstacle avoidance sub-agent.

| Trigger | Reward | Explanation |
| --- | --- | --- |
| Reach the target point. | +1 | Means that it has avoided every hazard. |
| Reach max step. | −0.5 | Aim: to avoid stand-by or loitering positions. |
| Colliding with another agent. | −0.5 | Aim: to avoid collisions. |
| Colliding with the edges of the environment. | −0.5 | |
| Colliding with an obstacle | −0.5 | |

## 4. Experiments and Results

This section first includes the description of the evaluation metrics to be used to analyze the swarm behavior. Then, it presents a series of evaluation scenarios and provides quantitative results for those metrics obtained through simulation.

### 4.1. Evaluation Metrics

The evaluation of the system's performance is achieved through several performance metrics computed from the corresponding data recorded during a simulation. It is important to note that these metrics reference the swarm's behavior, as opposed to the individual behaviors of the swarm's drones that make up the final group behavior. Hence, the complete swarming system described in this document is evaluated using the following metrics:

- Revisit period: This metric measures the frequency at which drones return to a specific area to update target information or detect changes in the area. This metric is computed by dividing the swarm's operation area in smaller cells forming a grid, and calculating for each cell the average duration of the periods when it is not visited by any UAV in the swarm. The goal of a swarm performing a search operation over an area should be to visit every cell as frequently as possible, minimizing this period.
- Search time: This metric refers to the total time it takes for the UAV swarm to find all the targets in the operation area. It is the minimum time for which the swarm controller has visibility of all the targets in the area at the same time.
- Percentage of detected targets: This metric assesses the success of the UAV swarm in detecting targets within the area. It is computed for each time instant, as the percentage of targets that are visible to the UAV swarm.
- Tracking continuity: This metric evaluates the ability of the UAV swarm to keep its targets within its field of view and obtain accurate information about their position and movement. It is computed as the ratio of time a target is in the swarm's field of view, in relation to the total tracking time for that target. The total tracking time for a certain target is the time from the first time it is detected and assigned to a UAV through a tracking task until the end of the operation or until the target exits the operation area, that is, the time where the swarm should keep that target in its field of view.
- Target acquisition time: This metric measures the time it takes for the UAV swarm to identify and start tracking a target after it has entered the surveillance area. If the search tasks are performed in an efficient way, this time should be minimized.

### 4.2. Simulation Scenarios and Results

The performance metrics described above are used to obtain results in an illustrative scenario. In it, the swarm aims to search and track targets in a squared operation area with sides of 6 km. Figures 8 and 9 show schematic 3D representations of the scenario, with targets, UAVs, and obstacles. These representations show the evolution of a scenario at two different times: once the initial deployment phase has been completed, and at a later stage. The main elements in this representation are:

- The targets are depicted as small vehicles on the ground. They are colored red and move with a constant speeds of 20 m/s, as described before.
- The UAVs are depicted as aircraft. UAVs' sensor coverage is depicted by drawing semi-spheres in blue, centered at the UAV's position and with a radius equivalent to the sensor coverage.
- Static obstacles are depicted as gray prisms, randomly distributed.
- Different colors on the ground are related to the definition of different search areas, to be assigned as tasks to different UAVs.
- Detection of a given target by a UAV sensor is depicted by joining both with a green line.

In the initial state (Figure 8), we may see that we have eight UAVs and eight search areas; no target has been acquired yet.

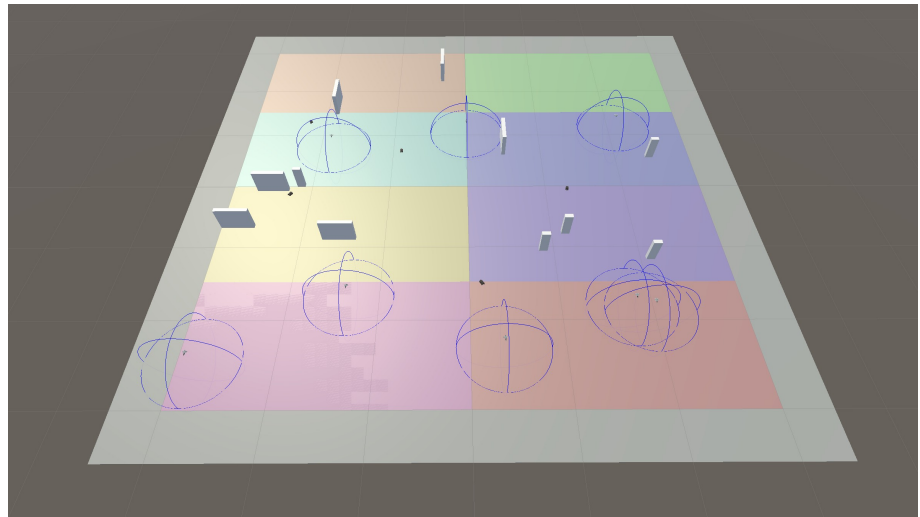

**Figure 8.** Mission beginning, after deployment (8 UAVs; 5 targets).

Later (Figure 9), we may see there are only five search areas, as three UAVs have been assigned tracking tasks (and they are actively detecting targets).

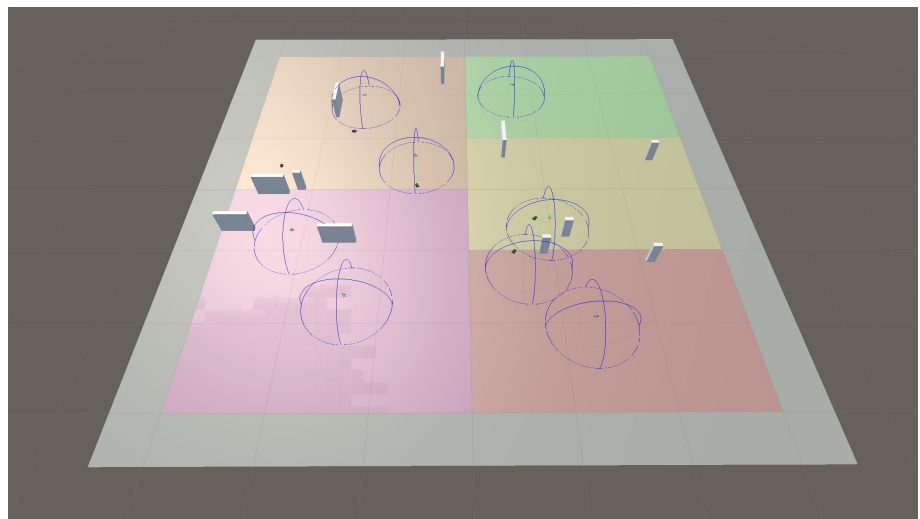

**Figure 9.** Mid-stage of mission (8 UAVs; 5 targets).

Based on variations in this baseline scenario (with varying numbers of targets and drones), several analyses have been performed:

1. A statistical analysis using 10 simulations to assess the behaviors of the UAV agents with the models obtained from the training process detailed in the previous section. This analysis, discussed in Section 4.2.1, uses a set number of drones and targets for all simulations.
2. An exploration on the swarm system's ability to adapt to a changing number of targets. For this, simulations with a varying number of targets are tested in Section 4.2.2.
3. An examination on how the system behaves with a varying number of drones within the swarm, as discussed in Section 4.2.3.

Overall, the results show that the proposed system is able to fulfill the surveillance mission and that it can correctly adapt to changing scenarios with varying numbers of UAVs and targets.

### 4.2.1. Statistical Analysis

This evaluation scenario consists of an eight-fixed-wing UAV swarm tracking five targets. The scenario has been run 10 times in parallel (shown in Figure 10) to perform a statistical analysis aggregating the swarm behavior on the random realizations of the scenario.

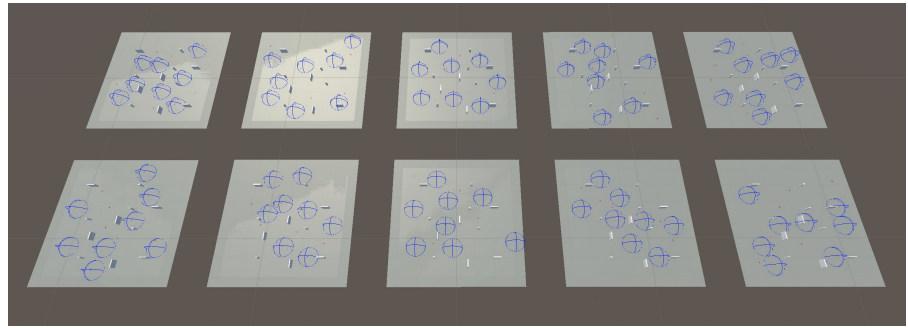

**Figure 10.** Test scenario.

The swarm's ability to search the operation area effectively can be seen in Figure 11, that represents a map of the operation area, divided by cells, with the mean of the revisit periods for all simulations. Additionally, the target acquisition time also serves to assess the swarm's search performance. The mean target acquisition times for the ordered targets (in order of identification) are shown in Table 5.

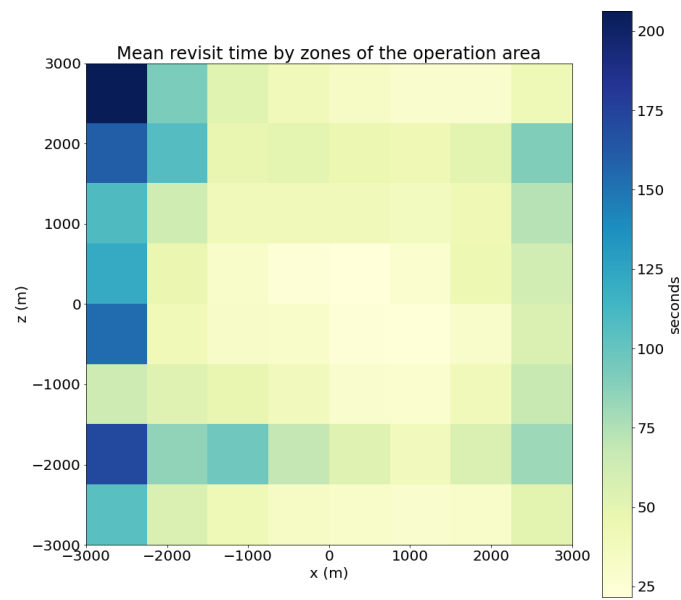

**Figure 11.** Mean revisit time by zone of the operation area throughout 10 simulations.

**Table 5.** Mean target acquisition time throughout 10 simulations.

|  | **First Target** | **Second Target** | **Third Target** | **Fourth Target** | **Fifth Target** |
|---|---|---|---|---|---|
| Target acquisition time | 2.8 s | 6.7 s | 19.1 s | 29.5 s | 71.9 s |

The map shows that the swarm is efficient in the exploration of the operation area, as most cells present a low revisit period. However, some areas on the borders of the operation area have higher revisit periods, suggesting that the swarm has more difficulty in exploring the edges of the operation area. In addition, from Table 5 it follows that the swarm acquires the targets in reasonable and efficient times given the size of the operation area.

On the other hand, the tracking continuity and the percentage of detected targets with time are useful to evaluate the swarm's tracking ability. The percentage of detected targets aggregated as the mean over the 10 simulations can be seen in Figure 12. Figure 13 represents the tracking continuity in a histogram, taking into account all the targets in the 10 simulations (a total of 50 targets), where the horizontal axis shows the percentage of tracking time in the swarm's field of view (since first detected), and the vertical axis represents the number of targets in each percentage segment.

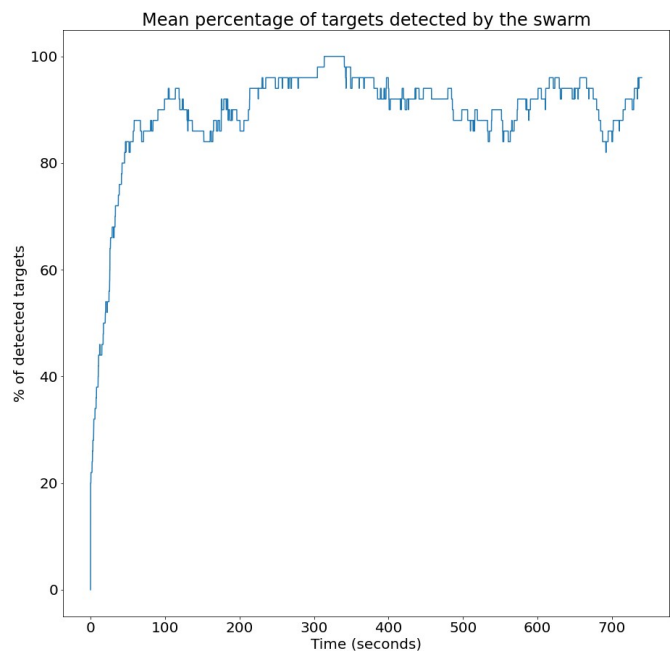

**Figure 12.** Mean percentage of targets detected by the swarm throughout 10 simulations.

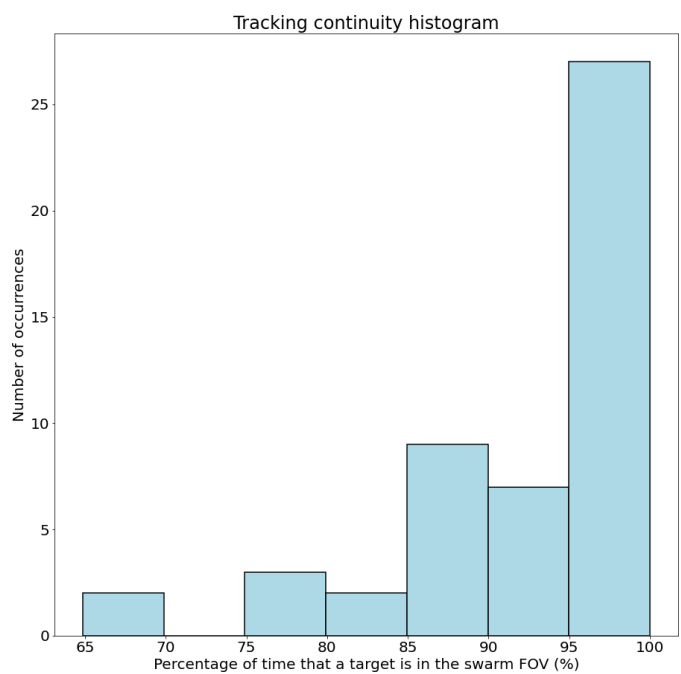

**Figure 13.** Histogram showing the percentage of tracking time that each target (50, 5 across 10 simulations) is in the swarm's field of view.

Although there are some targets that are lost by the swarm at some points in time, all targets are consistently tracked throughout the simulations, and target losses are rapidly recovered in general. In terms of tracking continuity, most of the targets are kept in the swarm's field of view for over 95% of the time since its first identification, which shows that the system tracks the targets consistently. The appearance of targets with lower percentages is in general due to occlusions behind obstacles.

4.2.2. Scenario with Changing Number of Targets

In addition to the previous statistical analysis, some modifications have been made to that scenario in order to show the adaptability of our system model to different conditions. The first of these modifications is related to the number of targets in the operation area, which has been incremented progressively (one, three, five, and six) and fixing the number of UAVs at eight. The results of these simulations (which also last 750 s) are shown in Figures 14 and 15.

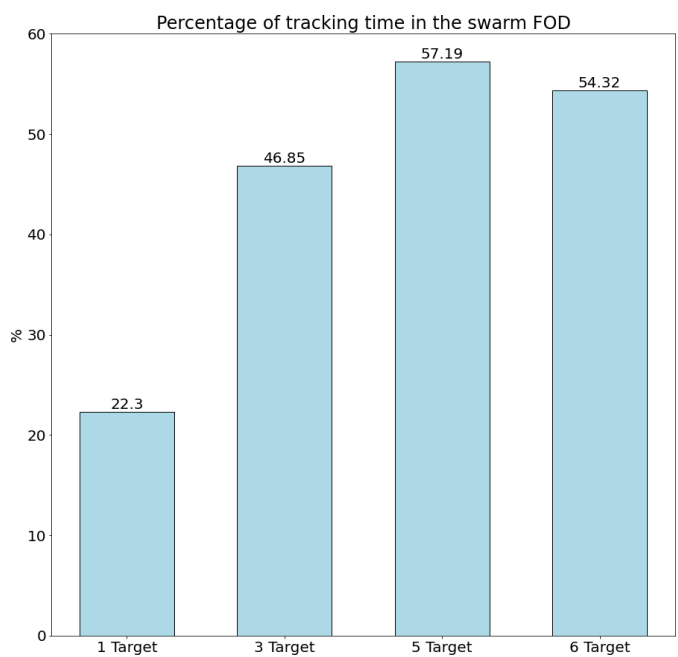

**Figure 14.** Revisit period and search time when modifying the number of targets.

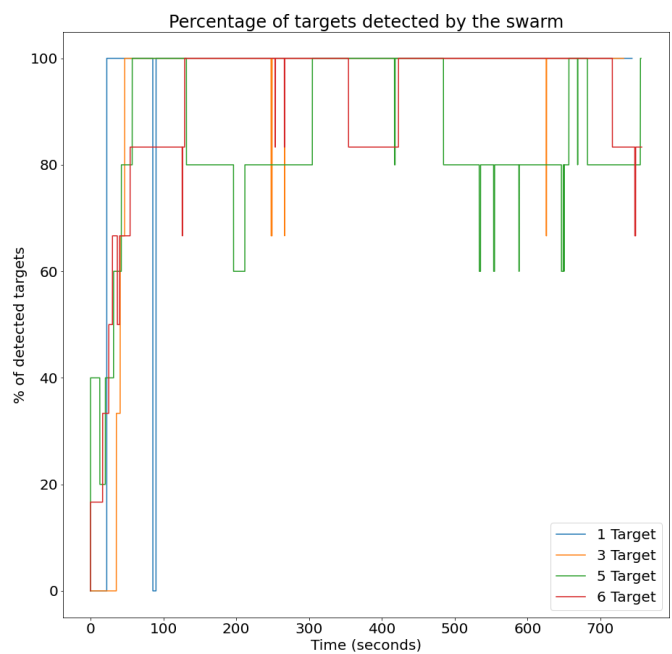

**Figure 15.** Percentage of targets detected by the swarm when modifying the number of targets.

It is inferred that the total search time generally increases with the number of targets. This is expected, as the swarm has to find an increasing number of targets. As for the percentage of detected targets, there are not so many differences between the four simulations since the tracking task is assigned to a single agent once the target is first detected. It is worth noting that in Figure 15 the blue line corresponds to the one-target simulation. Hence, the jump from 100% to 0% is just the loss of that single target; however, it is recovered a few seconds later.

In conclusion, the metric that is most affected is the search time, which make logical sense given that the number of agents is fixed and the number of targets is modified.

4.2.3. Scenario with Changing Number of Drones

The other scenario modification is changing the number of drones, which has been established at 8, 9, and 10. On this occasion, the number of targets is set at five. The results of these simulations are represented in Figures 16 and 17.

The former figure shows that the revisit time period improves significantly as the number of UAVs grows. From the latter, it is inferred that the greater the number of UAVs, the lower the ordered acquisition time. Both metrics provide the expected results.

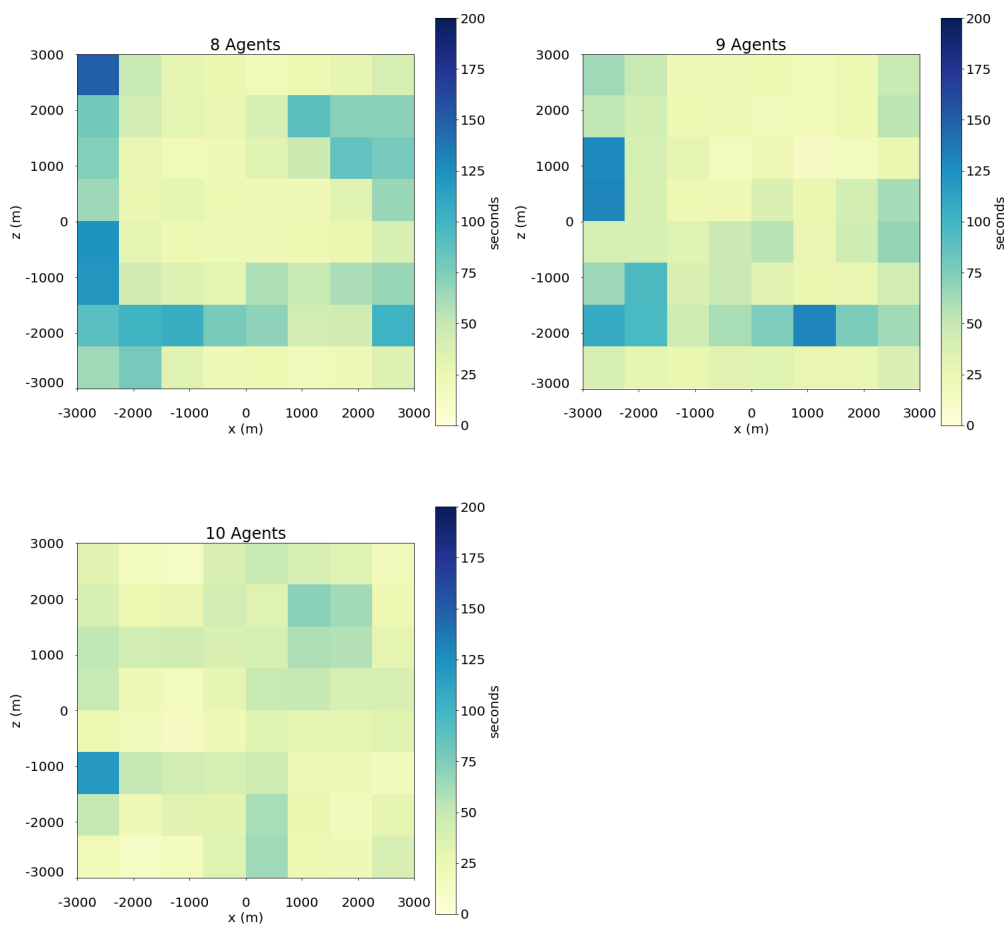

**Figure 16.** Mean revisit time by zone of the operation area when modifying the number of UAVs.

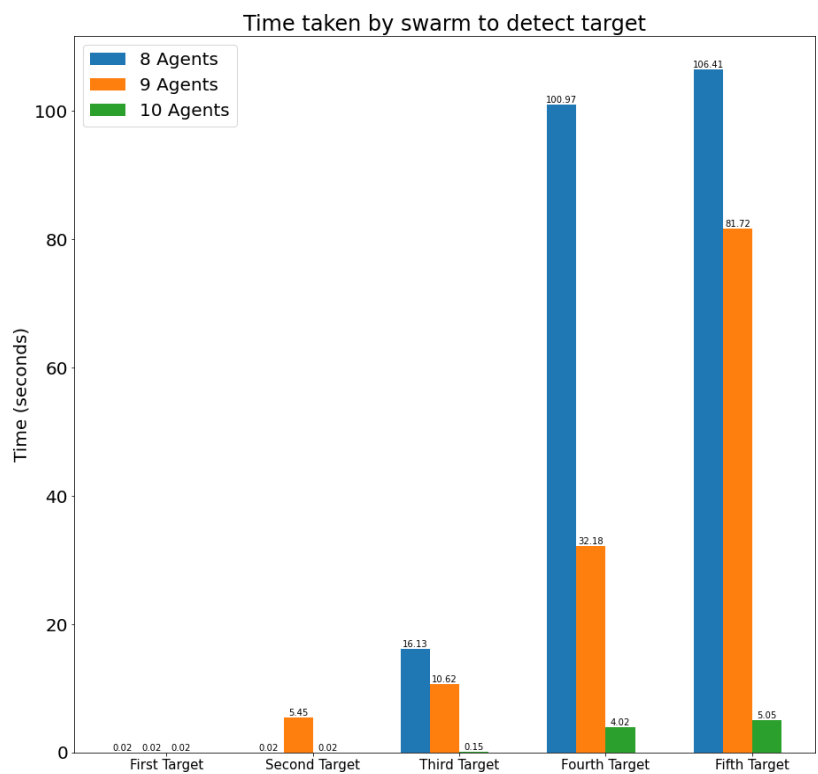

**Figure 17.** Target acquisition time when modifying the number of UAVs.

## 5. Conclusions

The system proposed in this paper uses hybrid AI incorporating DRL algorithms to obtain policies that may solve the whole surveillance problem (search, tracking, collision avoidance), taking into account obstacles in the environment, while other approaches in the literature using RL tend to neglect this aspect. In the current state of development, the performance of hybrid AI with DRL for the surveillance application seems correct, similar to that of traditional control methods in standard circumstances. This performance strongly depends on the fidelity to the mission of the training scenarios. A great effort is needed in the development of representative training situations. The creation of adequate action policies through reinforcement learning requires the careful design of the reward functions, which requires an in-depth study of each scenario and mission type. There is no clear method of determining these reward functions, and their determination depends on expert knowledge and experience. What makes the design process slow is needing to perform many design iterations, demanding significant computational resources. In any case, the obtained results show that the proposed divide-and-conquer methodology for the RL training, based on task-oriented DRL sub-agents, is effective in allowing a simplified training process to be performed.

Generating a representative collection of environment situations rich enough for the training of AI algorithms involves the use of realistic scenario simulators, as it is impossible and unsafe to implement all the hypothetical situations that can occur in a scenario from measurements in the real world.

From all the above, an important limitation of RL-based approaches is that they are expensive from the computational point of view, development time, and invested resources. It could be difficult to train a swarm control system with algorithms based on AI techniques for an imminent mission with a specific/new scenario.

Considering this limitation and others, the following lines of work can be suggested in the development of an operational prototype of swarm control based on reinforcement learning.

- Design of more realistic simulation environments that allow the efficient generation of training data for RL-based control algorithms, even tailored for a specific mission, in an acceptable time interval.
- Development of more efficient training systems for reinforcement learning. Both algorithmic and hardware aspects need to be addressed.
- Related to the previous point, a more systematic approach for the identification of the most relevant input features and their most compact representation, as well as of higher-level UAV control actions, might allow for more effective training processes.
- Further hybridization of classical and DRL approaches could be investigated, at all swarming layers.
- Development of structured design methods to obtain the actions sets and the reward functions for the swarm training phase adapted to a specific military mission.

**Author Contributions:** Conceptualization, R.A., G.d.M. and J.A.B.; funding acquisition, A.M.B. and J.A.B.; methodology, R.A., G.d.M. and J.A.B.; software, R.A. and D.C.; supervision, J.A.B. and A.M.B.; validation, R.A., D.C. and G.d.M.; writing—original draft, R.A.; writing—review and editing, J.A.B., D.C., G.d.M. and A.M.B. All authors have read and agreed to the published version of the manuscript.

**Funding:** This research was funded by MCIN/AEI/10.13039/501100011033 under Grant PID2020-118249RB-C21, and by the European Union's Horizon 2020 Research and Innovation Programme under Grant Agreement No 101021797.

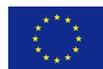

**Institutional Review Board Statement:** Not applicable.

**Informed Consent Statement:** Not applicable.

**Data Availability Statement:** Not applicable.

**Acknowledgments:** The authors acknowledge the initial work performed in the problem by Enrique Parro and Cesar Alberte, which allowed them to develop preliminary versions of the swarming training and demonstration infrastructure used for this research

**Conflicts of Interest:** The authors declare no conflict of interest.